\documentclass[runningheads]{llncs}
\usepackage{graphicx}
\usepackage[utf8]{inputenc} 
\usepackage[T1]{fontenc}    
\usepackage{hyperref}       
\usepackage{url}            
\usepackage{booktabs}       
\usepackage{amsfonts}       
\usepackage{nicefrac}       
\usepackage{microtype}      
\usepackage{graphicx}
\usepackage{xcolor}
\usepackage{grffile}
\usepackage{amsmath}
\usepackage{graphicx}
\usepackage{verbatim}
\usepackage{setspace}
\usepackage{subcaption}
\usepackage{verbatim}
\usepackage{multirow}
\usepackage{mwe}
\usepackage{verbatimbox}
\usepackage{array,booktabs}
\usepackage[inline]{enumitem}
\usepackage{mathtools}
\usepackage{floatrow}
\usepackage{multirow}
\usepackage{placeins}
\usepackage{array}
\usepackage{hypcap}
\usepackage{subcaption}
\usepackage{capt-of}
\usepackage{arydshln}
\usepackage{booktabs}
\usepackage{wasysym}
\usepackage{wrapfig}
\usepackage[ruled,linesnumbered,lined]{algorithm2e}
\usepackage{adjustbox}

\DeclareMathOperator*{\argmin}{arg\,min}

\newcommand{\spec}{I/O}

\newcommand{\ireng}{IReEn}

\begin{document}
\title{IReEn: Reverse-Engineering of Black-Box Functions via Iterative Neural Program Synthesis}
\titlerunning{IReEn}
%
\author{Hossein Hajipour\inst{1} \and
Mateusz Malinowski\inst{2}\and 
Mario Fritz\inst{1}}
\authorrunning{H. Hajipour et al.}
%
\institute{CISPA Helmholtz Center for Information Security \and
DeepMind
}
\maketitle              
\begin{abstract}
In this work, we investigate the problem of revealing the functionality of a black-box agent. Notably, we are interested in the interpretable and formal description of the behavior of such an agent. Ideally, this description would take the form of a program written in a high-level language. This task is also known as {\it reverse engineering} and plays a pivotal role in software engineering, computer security, but also most recently in interpretability. In contrast to prior work, we do not rely on privileged information on the black box, but rather investigate the problem under a weaker assumption of having only access to inputs and outputs of the program. 
We approach this problem by iteratively refining a candidate set using a generative neural program synthesis approach until we arrive at a functionally equivalent program. 
We assess the performance of our approach on the Karel dataset. Our results show that the proposed approach outperforms the state-of-the-art on this challenge by finding an approximately functional equivalent program in 78\% of cases -- even exceeding prior work that had privileged information on the black-box.
\end{abstract}

\section{Introduction}

\begin{wrapfigure}[11]{R}{0.5\linewidth}
\vspace{-.8cm}
\centering
    \includegraphics[width=0.99\linewidth]{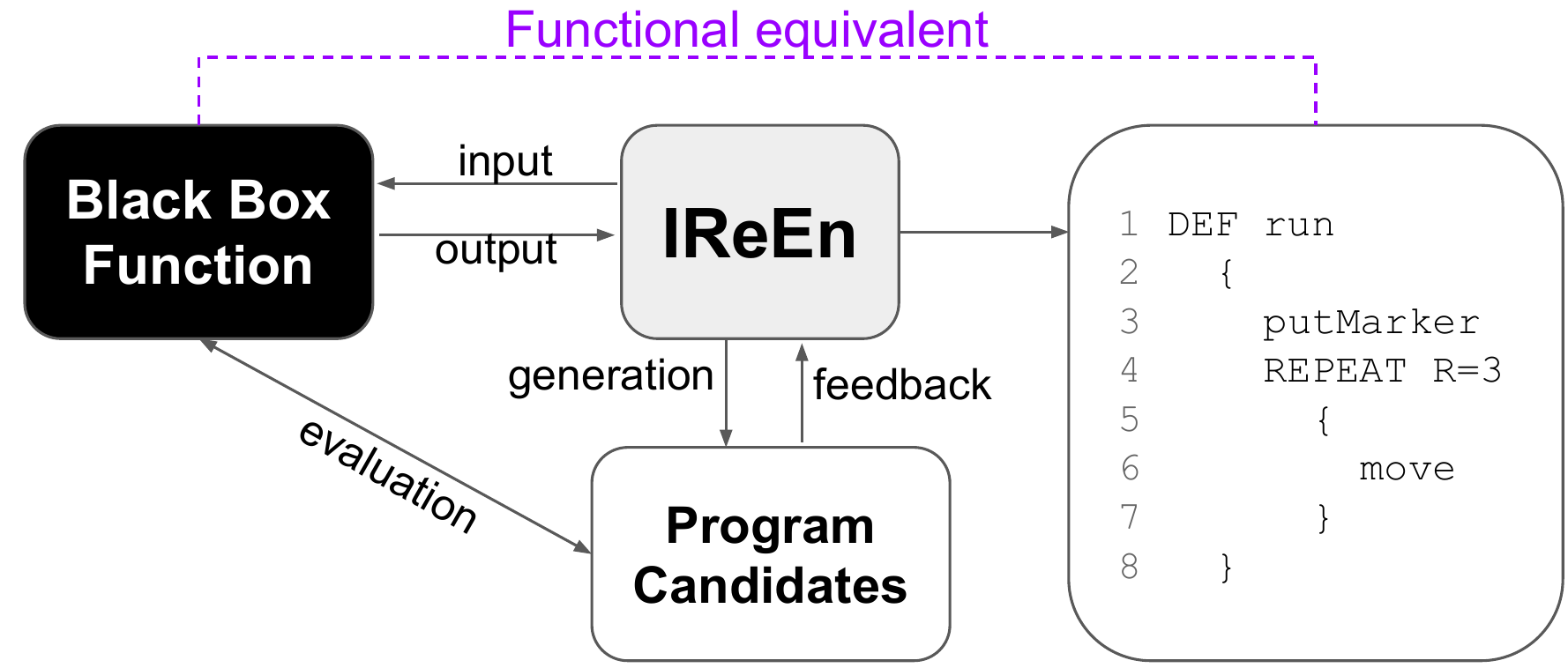}
    \caption{An example of revealing the functionality of a black-box function using only input-output interactions.}
\label{fig:teaser}
\end{wrapfigure}

Reverse-engineering (RE) 
is about
gaining insights into the inner workings of a mechanism, which often results in the capability of reproducing the associated functionality. 
In our work, we consider a program to be a black-box function that we have no insights into its internal mechanism, and we can only interface with it through inputs and the program generated outputs. This is a desired scenario in software engineering \cite{lee2011tie,fu2019coda}, or security, where we reverse-engineer, e.g., binary executables for analysis and for finding potential vulnerabilities \cite{kolbitsch2009effective,yakdan2016helping}. Similar principles have been applied in the program synthesis domain to understand the functionality of the given program \cite{jha2010oracle,solar2006combinatorial}. Furthermore, a similar paradigm is used to reverse-engineer the brain to advance knowledge in various brain-related disciplines \cite{markram2012human} or seeking interpretation for reinforcement learning agents \cite{verma2018programmatically}.

Despite all of the progress in reverse-engineering the software and machine learning models, there are typical limitations in the proposed works. E.g., in the decompilation task, one of the assumptions is to have access to the assembly code of the black-box programs or other privileged information, which is a significant information leak about the black-box function. In program synthesis, a common issue is that the problem is considered in a relaxed setting, where they only synthesize loop-free or conditional-free programming languages \cite{jha2010oracle}. Furthermore, in deep learning, a common issue is that the reverse-engineered models usually are not represented in an interpretable and human-readable form. 

In recent work, neural program synthesizers are employed to recover a functional and interpretable form of a black-box program that is generated based only on I/Os examples \cite{bunel2018leveraging}.  On close inspection, however, it turns out, that these approaches also leverage privileged information, by relying on a biased sampling strategy of I/Os that was obtained under the knowledge of the black-box function.

In contrast to prior work, we propose an iterative neural program synthesis scheme which is the first to tackle the task of reveres-engineering in a black-box setting without any access to privileged information.
Despite the weaker assumptions and hence the possibility to use our method broadly in other fields, we show that in many cases it is possible to reverse engineer approximately functionally equivalent programs on the Karel dataset benchmark. We even achieve better results than prior work that has access to privileged information.

We achieve this by an iterative reverse-engineering approach. We query a black-box function using random inputs to obtain a set of I/Os, and refine the candidate set by an iterative neural program synthesis scheme. This neural program synthesis is trained with pairs of I/Os and target programs. To adapt our program synthesize to the domain of random I/Os we fine-tune our neural program synthesize using random I/Os and the corresponding target program.

To summarize the contributions of this work are as follow:
\begin{enumerate}
    \item We propose an iterative neural program synthesizer scheme to reverse-engineer a functionally equivalent form of the black-box program. To the best of our knowledge, this is the first approach that operates in a black-box setting without privileged information.
    \item We proposed functional equivalence metric in order to quantify progress on this challenging task.
    \item We evaluate our approach on Karel dataset, where our approach successfully revealed the underlying programs of 78\% of the black-box programs. Our approach outperforms prior work despite having access to less information due to weaker assumptions.
\end{enumerate}

\section{Related Work}

\paragraph{Reverse-Engineering of programs.}
Decompilation is the task of translating a low-level program into a human-readable high-level language. Phoenix \cite{brumley2013native} and Hex-Rays \cite{Hexray} are conventional decompilers. These decompilers are relying on pattern matching and hand-crafted rules, and often fail to decompile non-trivial codes with conditional structures. Fu et al. \cite{fu2019coda} proposed a deep-learning-based approach to decompile the low-level codes in an end-to-end fashion. In the decompilation task, the main assumption is to have access to a low-level code of the program. However, in our approach, our goal is to represent a black-box function in a high-level program language only by relying on input-output interactions.   

\paragraph{Reverse-Engineering of neural networks.} Reverse-engineering neural network recently has gained popularity. Oh et al. \cite{oh2019towards} proposed a meta-model to predict the attributes of the black-box neural network models, such as architecture and optimization process. Orekondy et al. \cite{orekondy2019knockoff} investigate how to steal the functionality of the black-box model only based on image query interactions. While these works try to duplicate the functionality of a black-box function, in this work our goal is to represent the functionality of the black-box function in a human-readable program language.

\paragraph{Reverse-Engineering for interpretability.}
In another line of work, Verma et al. \cite{verma2018programmatically} and  Bastani et al. \cite{bastani2018verifiable} proposed different approaches to have interpretable and verifiable reinforcement learning. Verma et al. \cite{verma2018programmatically} designed a reinforcement learning framework to represent the policy network using human-readable domain-specific language, and Bastani et al. \cite{bastani2018verifiable} represent policy network by a training decision tree. Both of these works are designed for a small set of RL problems with a simple program structure. However, in our work, we consider reverse-engineer a wide range of programs with complex structures.

\paragraph{Program synthesis.}
Program synthesis is a classic task which has been studied since the early days of Artificial Intelligence \cite{waldinger1969prow,manna1975knowledge}. Recently there has been a lot of recent progress in employing the neural-networks-based approaches to do the task of program synthesis. One type of these approaches called \textit{neural program induction} involves learning a machine learning model to mimic the behavior of the target program \cite{graves2014neural,johnson2017inferring,devlin2017neural}. Another type of approach is \textit{neural program synthesis}, where the goal is to learn to generate an explicit discrete program in a domain-specific program language. Devlin
et al. \cite{devlin2017robustfill} proposed an encoder-decoder neural network style to learn to synthesize programs from input-output examples. Bunel et al. \cite{bunel2018leveraging} synthesizing Karel programs from examples, where they learn to generate program using a deep-learning-based model by leveraging the syntax constraints and reinforcement learning. Shin et al. \cite{shin_imporivng} and Chen et al. \cite{chen2018execution} leverage the semantic information of execution trace of the programs to generate more accurate programs. These works assume that they have access to the crafted I/O examples. However, in this work, we proposed an iterative program synthesis scheme to deal with the task of black-box program synthesis, where we only have access to the random I/O examples.

\section{Problem Overview}
\label{sec:prob}
In this section, we formulate the problem description and our method. We base our notation on~\cite{bunel2018leveraging,shin_imporivng,chen2018execution}.

\paragraph{Program synthesis.}
\label{subsec:progformula}
Program synthesis deals with the problem of deriving a program in a specified programming language that satisfies the given specification. We treat input-output pairs $\spec = \left\{(I^k,O^{k})\right\}_{k=1}^K$ as a form of specifying the functionality of the program. This problem can be formalized as finding a solution to the following optimization problem:
\begin{eqnarray}
\label{eq:problem_synthesis}
    \argmin_{p \in \mathcal{P}} & \Omega(p) &\\
    \text{s.t.} & p(I^k) = O^k &\forall k\in\{1,\ldots,K\}
\end{eqnarray}
where $\mathcal{P}$ is the space of all possible programs written in the given language, and $\Omega$ is some measure of the program. For instance, $\Omega$ can be a cost function that chooses the shortest program.

\begin{wrapfigure}[13]{R}{0.55\linewidth}
\vspace{-.7cm}
    \centering
    \includegraphics[width=0.96\linewidth]{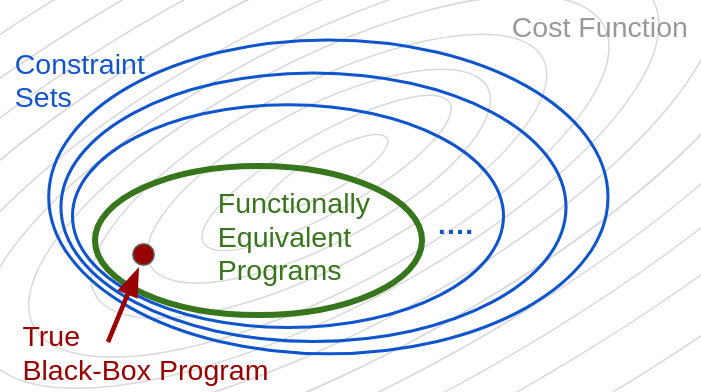}
    \caption{Illustration of the optimization problem, functional equivalence, and feasible sets w.r.t. nested constraint sets.}
    \label{fig:constraints}
\end{wrapfigure}

The situation is illustrated in \autoref{fig:constraints}. For many applications -- also the one we are interested in -- there is a true underlying black-box program that satisfies all the input-output pairs. As most practical languages do not have a unique representation for certain functionality or behavior, a certain set of functionally equivalent programs will remain indistinguishable even given an arbitrary large number of input-output observations and respective constraints in our optimization problem. Naturally, by adding more constraints, we obtain a nested constraint set that converges towards the feasible set of functionally equivalent programs.

\paragraph{Program synthesis with privileged information.} Recent works implicitly or explicitly incorporates insider information on the function to reverse-engineer. This can come in the form of a binary of the compiled code or an informed sampling strategy of the input-output pairs. 
It turns out that  the 
majority of recent research implicitly uses privileged information via biased sampling scheme in terms of \emph{crafted} specifications~\cite{bunel2018leveraging,chen2018execution,pattis1981karel,shin_imporivng}. Note that in order to arrive at these specifications, one has to have access to the program $P$ under the question as they are designed to capture e.g. all branches of the program. We call these crafted specifications \emph{crafted I/Os} and will investigate later in detail how much information they leak about the black-box program.

\paragraph{Black-Box program synthesis.}
\label{subsec:blackformula}
In our work, we focus on a black-box setting, where no such side or privileged information is available.
Hence, we will have to defer initially to randomly generate $K$ inputs $\{I^k\}_{k=1}^K$ and next query the program $p$ to obtain the corresponding outputs $\{O^k\}_{k=1}^K$. 
Such generated input-output pairs become our specification that we use to synthesize programs. Note that, unlike the previous setting, here we take advantage of querying the black-box program $p$ in an active way, even though the whole procedure remains automatic. To generate random inputs we follow the procedure proposed by Bunel et al. \cite{bunel2018leveraging}. We call the obtained I/Os in the black-box setting \emph{random I/Os}. It turns out (as we will also show in our experiments), that indeed such random, uninformed input queries yield significantly less information than the \emph{crafted I/Os} used in prior work. Hence, to arrive at an effective and black-box approach, in the following we propose an iterative reverse-engineering scheme, that gradually queries more relevant inputs. 

\section{IReEn: Iterative Reverse-Engineering of Black-Box Functions}
\label{sec:itr}

Reverse-engineering a black-box function and representing it in a high-level language is a challenging task. The main reason is that we can only interact with the black-box function using input-output examples.  In addition, solving the above constraint optimization problem is intractable. Therefore, in the following, we relax the optimization problem to a Bayesian inference problem and show how to iteratively incorporate additional constraints in order to arrive at a functional equivalent program with respect to the black-box function.

\begin{figure}[t] 
	\centering
	    \centering
		\includegraphics[width = 1.0\textwidth]{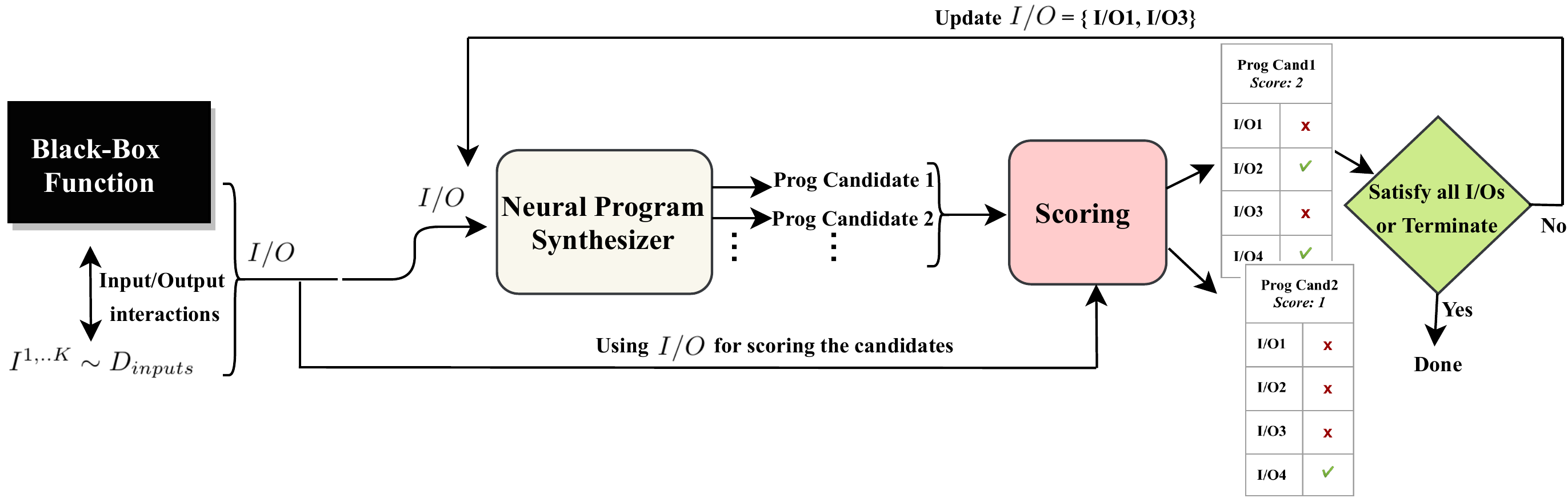}
	
	\caption{Overview of the proposed iterative neural program synthesis approach.}
	\label{fig:pipeline}
	\vspace{-.1cm}
\end{figure}

\autoref{fig:pipeline} provides an overview of our iterative neural program synthesis scheme to reverse-engineer the given black-box function. In the first step, we obtain the I/Os by querying the black-box function using random inputs drawn from a distribution of inputs. We condition the neural program synthesizer on the obtained I/Os. Neural program synthesizer outputs the potential program candidate(s), and then we use a scoring system to score the generated candidates. For example, in this figure "program candidate 1" satisfied two out of four sample I/Os, so its score will be 2. If the best candidate does not cover all of the I/Os, we select a subset of I/Os which were not covered by the best candidate program to condition them on the program synthesizer for the next iteration.

\subsection{Finding Programs given Input-Output Constraints}
\label{subsec:bb_syn}

Even for a small set of input-output constraints, finding the feasible set of programs that satisfies these I/Os is not tractable due to the discrete and compositional nature of programs. We approach this challenging problem by relaxing the constraint optimization problem to a Bayesian Inference problem. 
In this way, samples of the model are solutions to the constraint optimization problem. In order to train such a generative model, we directly optimize the Neural Program Synthesis approach based on Bunel et al. \cite{bunel2018leveraging}. This is a conditional generative model that samples candidate programs by conditioning on the input-output information.

\begin{align}\label{eq:syn_black_box} 
\hat{P} \sim \Psi(I/O).
\end{align}
Where $\hat{P}$ is a set of sampled solutions that are program candidate(s) $\{\hat{p}_1,...,\hat{p}_C\} \in \hat{P}$ and $C \geq 1$.

In detail, we train a recurrent encoder and decoder for program synthesis on a set of ground-truth programs $\{p_i\}_i$ and specifications $\{I/O_i\}_i$.
Each specification is a set of $K$ pairs $\spec_i = \{(I_i^k,O_i^k)\}_{k=1}^{K}$ where the program needs to be consistent with, that is, $p_i(I_i^k) = O_i^k$ for all $k\in\{1,\ldots,K\}$. In our work, we pre-train the program synthesis proposed by Bunel et al. \cite{bunel2018leveraging}, where they use encoder-decoder neural networks to generate the desired program given input-output specifications. Note that the synthesizer is dependent on the input specification, that is, different $\spec_a$ and $\spec_b$ may produce different programs through the synthesis, i.e., $\Psi(\spec_a) = p_a$ and $\Psi(\spec_b) = p_b$. For a detailed discussion, e.g. of the I/O encodings, we refer to Bunel et al. \cite{bunel2018leveraging}.

\subsection{Sample Rejection Strategy}
Naturally, we expect approximation errors of the optimization problem by the generative model. Two main sources of error are (1) challenges to approximate the discontinuous target distribution (2) only a limited number of constraints can be incorporated in the conditional generative model.
In order to correct for these errors, we follow up with a sample rejection stage 
based on a scoring of the  generated program candidates. We use random I/Os obtained from interacting with black-box function to evaluate the generated programs, and score them based on the number of the I/Os which were covered by the programs (See \autoref{fig:pipeline}).  
While in principle, any failed I/O should lead to rejecting a candidate, empirically, we find that keeping the highest scoring samples turns out to be advantageous and prevents situations where no candidates would remain.
\subsection{Iterative Refinement}
We are still facing two major issues: (1) As we have motivated before and also our experiments will show, querying for certain I/O pairs is more informative than others. Hence, we seek an iterative approach that yields more informative queries to the black box. (2) Due to the computational bottleneck, the conditional generative model only takes a small number of constraints, while it is unclear which constraints to use in order to arrive at the ``functional equivalent'' feasible set.

Similar problems have been encountered in constraint optimization, where {\it column generation algorithms / delayed constraint generation} techniques  have been employed to deal with large number of constraints     \cite{ford1958suggested}. Motivated by these ideas, we propose an iterative strategy, where we condition in each step on a set of violated constraints that we find.

In detail, we present the algorithm of the proposed method in Algorithm \autoref{alg:1}. Iterative synthesis function takes synthesizer $\Psi$ and a set of I/Os (line 1). In line 2 we initial the $s_\textit{best}$ to zero. Note that we use $p_\textit{best}$ to store the best candidate, and $s_\textit{best}$ to store the score of the best candidate. In the iterative loop, we first condition the program synthesizer on the given $I/O$ set to get the program candidates $\hat{P}$ (line 5). Then we call \textit{Scoring} function to score the program candidates in line 6. The scoring function returns the best program candidate, the score of that candidate, and the new set of $I/O$ where the new I/Os are the one which were not satisfied by $\hat{p}_\textit{best}$. Note that $\hat{p}_\textit{best}$, and $\hat{s}_\textit{best}$ store the best candidate and the score of it for the current iteration. Then at line 7, we check if $\hat{s}_\textit{best}$ for the current iteration is larger than the global score $s_{best}$, if the condition satisfies we update the global $p_\textit{best}$, and $s_\textit{best}$ (line 8-9). In line 12 we return the best candidate $p_\textit{best}$ after searching for it for $n$ iterations.

\begin{wrapfigure}[15]{R}{0.60\linewidth}
\vspace{-0.1cm}
\begin{minipage}{0.98\linewidth}
\begin{algorithm}[H]
\label{alg:1}
\DontPrintSemicolon
\SetAlgoLined
\SetKwFunction{FMain}{IterativeSynthesis}
\SetKwProg{Fn}{Function}{:}{}
\Fn{\FMain{$\Psi$, $I/O$}}{

$s_{best}$ = 0 \tcp{To keep the best score.}
n = \textit{constant} \,\tcp{e.g., n=10}

\For{$i\gets1$ \KwTo $n$}{
  $\hat{P}$ = $\Psi(I/O)$\;
  $\hat{p}_\textit{best}$, $\hat{s}_\textit{best}$, $I/O$ = $\textit{Scoring}(\hat{P})$ \;
  \If{$s_\textit{best}$ < $\hat{s}_\textit{best}$}{
  $p_\textit{best}$ = $\hat{p}_\textit{best}$\;
  $s_\textit{best}$ = $\hat{s}_\textit{best}$\;
   }
  }
  \textbf{return} $p_\textit{best}$\;
  
  }
\textbf{End Function}
 \caption{Iterative Algorithms}
\end{algorithm}
\end{minipage}
\end{wrapfigure}

\subsection{Fine-tuning}
\label{subsec:ft}
The goal of synthesizer $\Psi$ is to generate a program for the given I/Os, so it is not desirable to generate a program that contains not-used statements (e.g. a while statement which never hit by the given I/Os). However, in the black-box setting, we only have access to the random I/Os, and there is no guarantee if these I/Os represent all details of the black-box program. So the synthesizer might need to generate a statement in the program which was not represented in the given I/Os. The question is how we can have a synthesizer that makes a balance between these two contradictory situations. To address this issue, we first train synthesizer $\Psi$ on the crafted I/Os and then fine-tune it on the random I/Os. Please note that we only use the crafted I/Os during training. We get the data for fine-tuning by pairing random I/Os with the target programs. We empirically find that fine-tuning the synthesizer can lead to better performance than training it using only random I/Os.

\section{Experiments}
In this section, we show the effectiveness of our proposed approach for the task of black-box program synthesis. We consider Karel dataset \cite{devlin2017neural,bunel2018leveraging} in a strict black-box setting, where we can only have access to I/Os by querying the black-box functions without any privileged information or informed sampling scheme. 
\subsection{The Karel Task and Dataset}
\begin{figure}[h] 
	\centering
	    \centering
		\includegraphics[width = 1.0\textwidth]{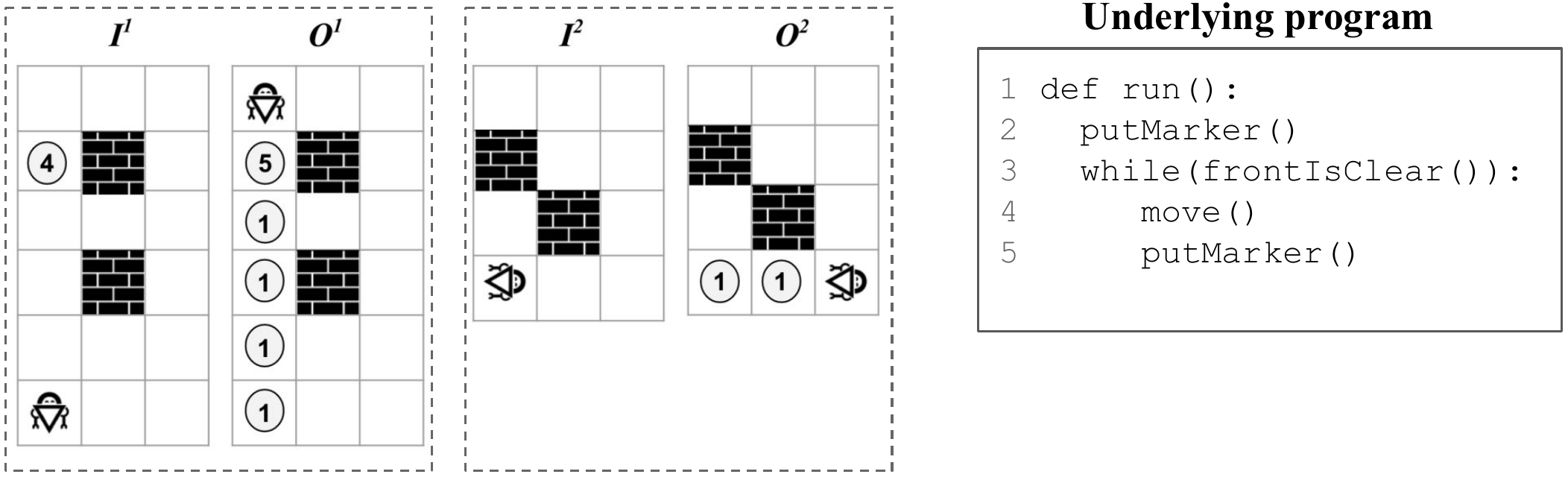}
	
	\caption{Example of two I/Os of a Karel task with the corresponding underlying program. The robot is Karel, the brick walls represent obstacles, and markers are represented with circles.}
	\label{fig:sup:karel_example}
	\vspace{-.1cm}
\end{figure}

To evaluate our proposed approach, we consider Karel programming language. 
Karel featured a robot agent in a grid world, where this robot can move inside the grid world and modify the state of the world using a set of predefined functions and control flow structures. 
Recently it has been used as a benchmark in several neural program synthesis works \cite{bunel2018leveraging,shin_imporivng,chen2018execution}. \autoref{fig:karelDSL} shows the grammar specification of this programming language \cite{bunel2018leveraging}. Using control flow structures such as condition and loop in the grammar of Karel makes this DSL a challenging language for the task of program synthesis. \autoref{fig:sup:karel_example} demonstrates an example of the Karel task with two I/O examples and the corresponding program.

Bunel et al. \cite{bunel2018leveraging} defined a dataset to train and evaluate neural program synthesis approaches by randomly sampling programs from the Karel's DSL. In this dataset, for each program, there is 5 I/Os as specification, and one is the held-out test sample. In this work, we consider the Karel's programs as black-box agent's task, and our goal is to reveal the underlying functionality of this black-box function by solely using input-output interactions. This dataset contains 1,116,854 pairs of I/Os and programs, 2,500 for validations, and 2,500 for testing the models. Note that, to fine-tune the synthesizer to the domain of random I/Os, we used 100,000 pairs of random I/Os and target program for training and 2,500 for validation.

\begin{figure}[htp]
  \begin{center}
    \begin{eqnarray*}
      \mbox{Prog } p & := & \mathtt{def }\ \mathtt{run}()\;:\ s \\
      \mbox{Stmt } s & := & \mathtt{while}(b): s \; |\  \mathtt{repeat}(r): s \;
                            |\ s_1; s_2 \; | \; a \\
                     & | & \mathtt{if}(b): s \; | \; \mathtt{ifelse}(b): s_1\ \mathtt{else}: s_2 \\
      \mbox{Cond } b & := & \mathtt{frontIsClear()} \; | \; \mathtt{leftIsClear()} \; | \; \mathtt{rightIsClear()} \\
                     & | & \mathtt{markersPresent()} \; | \;
                           \mathtt{noMarkersPresent()} \ | \ \; \mathtt{not} \; b  \\
      \mbox{Action } a & := & \mathtt{move()} \; | \; \mathtt{turnRight()} \; | \; \mathtt{turnLeft()} \\
                     & | & \mathtt{pickMarker()} \; | \; \mathtt{putMarker()} \\
      \mbox{Cste } r & := & 0 \; |\ 1 \; | \ \dotsb \; |\  19
    \end{eqnarray*}
  \end{center}
  \caption{The grammar for the Karel programming language.}
  \label{fig:karelDSL}
\end{figure}

\subsection{Training and Inference}
We train the neural program synthesizer using the Karel Dataset. To train this synthesizer we employ the neural networks architecture proposed by Bunel et al. \cite{bunel2018leveraging}, and use that in our iterative refinement approach as the synthesizer. Note that, to fine-tune the synthesizer model on random I/Os we use Adam optimizer \cite{kingma15adam} and the learning rate $10^{-5}$. We fine-tune the synthesizer model for 10 epochs. During inference, we use beam search algorithm with beam width 64 and select top-k program candidates.
\subsection{Functional Equivalence Metric}
\label{subsec:functional}
In \cite{bunel2018leveraging,shin_imporivng},  two metrics have been used to evaluate the trained neural program synthesizer. 
\begin{enumerate*}
    \item Exact Match: A predicted program is an exact match of the target if it is the same as the target program in terms of tokens.
    \item Generalization: A predicted program is a generalization of the target if it satisfies the I/Os of the specification set and the held-out example.
    
\end{enumerate*}
Both of these metrics have some drawbacks. A predicted program might be functionally equivalent to the target program but not be the exact match. On the other side, a program can be a generalization of the target program by satisfying a small set of I/Os (in Bunel et al. \cite{bunel2018leveraging} 5 I/Os has been used as specification and 1 I/O is considered as held-out). However, it might not cover a larger set of I/Os for that target program. To overcome this issue, in this work we proposed the Functional Equivalence metric, where we consider a predicted program as an approximately functional equivalent to the target program if it covers a large set of I/Os which were not been used as the specification in the synthesizing time. To get the set of I/Os, we generate the inputs randomly and query the program to get the outputs. We check if these inputs hit all of the branches of the target program. In our experiments, we found that with using more I/Os the more predicted programs we discover to not be functionally equivalent to the target programs. We found 100 number of I/Os as a point where the number of approximately functional equivalent programs stay stable in the evaluations.

\subsection{Evaluation}

\begin{table}[]

\fontsize{8.5}{10.5}\selectfont
\begin{center}
\caption{Top: Results of performance comparison of our approach in different settings using random I/Os for black-box program synthesis. Random I/Os mean that we use randomly obtained I/Os in the black-box setting, FT refers to fine-tuned model, and \ireng\, denote to our iterative approach. Bottom: Results of Bunel et al. \cite{bunel2018leveraging} when we use crafted I/Os. top-1 denote the results for the most likely candidate, and top-50 denote the results for 50 most likely candidates.}
\label{table:acc}
\end{center}

\begin{center}
\begin{tabular}{llllllllll}
\toprule
Models       & \multicolumn{2}{c}{Generalization }& \multicolumn{2}{c}{Functional} & \multicolumn{2}{c}{Exact Match} \\ \cmidrule(lr){1-1}\cmidrule(lr){2-3}\cmidrule(lr){4-5}\cmidrule(lr){6-7}
             & top-1 & top-50 & top-1 & top-50  & top-1 & top-50   \\
             \cmidrule(lr){2-7}
Random I/Os &   57.12\% &   71.48\%       & 49.36\%         &  63.72\%        &     34.96\%       &    40.92\%    \\ 

Random I/Os + FT &  64.72\%       &   77.64\%     &   55.64\%     &    70.12\%      &    39.44\%      &    45.4\%     \\ 
Random I/Os + \,\ireng    &    76.20\%     &  85.28\%          &  61.64\%        &   73.24\%        &     40.95\%   &    44.99\%      \\ 
Random I/Os + FT + \,\ireng & \textbf{78.96\%}        &    \textbf{88.39}\%         &   \textbf{65.55\%}       &  \textbf{78.08\%}       &   \textbf{44.51\%}     &  \textbf{48.11\%} \\

\cmidrule(lr){1-7}
Crafted I/Os (\cite{bunel2018leveraging}) &    73.12\%     &  86.28\%          &  55.04\%        &   68.72\%        &     40.08\%   &    43.08\%      \\ 

\\\bottomrule
\end{tabular}
\end{center}
\end{table}

We investigate the performance of our approach in different settings to do the task of black-box program synthesis. 
To evaluate our approach we query each black-box program in the test set with 50 valid inputs to get the corresponding outputs. Using the obtained 50 I/Os we synthesize the target program, where we use 5 out of 50 I/Os to conditions on the synthesizer and use 50 I/Os to score the generated candidate and find the best one based on sample rejection strategy.  In our iterative approach in each iteration, using sample rejection strategy we find a new 5 I/Os among the 50 I/Os to condition on the synthesizer for the next iteration. To evaluate the generated programs, in addition to generalization and exact match accuracy, we also consider our proposed metric called Functional Equivalence (\autoref{subsec:functional}). To compute the functional equivalency we use 100 I/Os which were not seen by the model. If the generated program satisfies all of 100 I/Os we consider it as a program which approximately functional equivalent to the target program. In all of the results top-k means that we use the given I/Os to find the best candidate among the "k" top candidates. 
For computing the results for all of the metrics we evaluate the best candidate among top-k.

\begin{table}[]

\fontsize{8.5}{10.5}\selectfont
\begin{center}
\caption{Top: Functional equivalence results of our approaches in synthesizing black-box programs with different complexity. Random I/Os means that we use randomly obtained I/Os in the black-box setting, FT refers to fine-tuned model, and \ireng\, denotes to our iterative approach. Bottom: Results of Bunel et al. \cite{bunel2018leveraging} when we use crafted I/Os. Action refers to programs that only contain action functions, Repeat denotes programs with action functions and only a repeat structure, While denotes programs with action functions and only a while control flow, If refers to the programs with action functions and only an if control flow, and Mix denotes programs with more than one control flow structures and action functions. top-1 denotes the results for the most likely candidate, and top-50 denotes the results for 50 most likely candidates.}
\label{table:control}
\end{center}
\begin{adjustbox}{width=\columnwidth,center}
\begin{tabular}{lllllllllll}
\toprule
Models       & \multicolumn{2}{c}{Action }& \multicolumn{2}{c}{Repeat} & \multicolumn{2}{c}{While} & \multicolumn{2}{c}{If} & \multicolumn{2}{c}{Mix}\\ \cmidrule(lr){1-1}\cmidrule(lr){2-3}\cmidrule(lr){4-5}\cmidrule(lr){6-7} \cmidrule(lr){8-9}  \cmidrule(lr){10-11}
             & top-1 & top-50 & top-1 & top-50  & top-1 & top-50 & top-1 & top-50 & top-1 & top-50 \\
             \cmidrule(lr){2-11}
Random I/Os &   95.59\% &   99.69\%       & 85.52\%         &  91.44\%        &     26.98\%       &    61.58\%    & 48.88\% & 72.69\% & 10.69\% & 27.12\%\\ 

Random I/Os + FT &  99.39\% &   99.76\%       & 90.72\%         &  96.38\%        &     56.50\%       &    82.22\%    & 52.06\% & 77.46\% & 14.33\% & 32.19\%     \\ 
Random I/Os +~\ireng    &    \textbf{99.84}\% &   99.84\%       & \textbf{96.38}\%         &  97.36\%        &     60.95\%       &    84.76\%    & 81.26\% & 89.84\% & 27.67\% & 49.94\%      \\ 
Random I/Os + FT +~\ireng & \textbf{99.84}\% &   \textbf{100}\%       & 95.39\%         &  \textbf{99.64}\%        &     \textbf{81.58}\%       &    \textbf{93.33}\%    & \textbf{81.52}\% & \textbf{92.06}\% & \textbf{32.08}\% & \textbf{56.22}\% \\

\cmidrule(lr){1-11}
Crafted I/Os &    99.08\% &   100.0\%       & 91.11\%         &  96.71\%        &     54.28\%       &    84.12\%    & 49.20\% & 79.68\% & 14.88\% & 33.84\%      \\ 

\\\bottomrule
\end{tabular}

\end{adjustbox}
\end{table}

\paragraph{Comparison to baseline and ablation.} \autoref{table:acc} shows the performance of our approach in different settings in the top, and the results of the neural program synthesizer proposed by Bunel et al. \cite{bunel2018leveraging} in the bottom. These results show that when we only use random I/Os (first row), there is a huge drop in the accuracy in all of the metrics in comparison to the results of crafted I/Os. However, when we fine-tune the synthesizer the results improve in all of the metrics, especially for the top-1 and top-50 functional equivalence accuracy. Furthermore, when we use our iterative approach for 10 iterations with the fine-tuned model (fourth row), we can see that our approach outperforms even the crafted I/Os in all of the metrics. For example, it outperforms crafted I/Os in functional equivalence and exact match metric by a large margin, 9\%, and 5\% respectively for top-50 results. 
\paragraph{Importance of the crafted I/Os.}
In \autoref{table:acc} in the top first row (Random I/Os) we use random I/Os to condition on the synthesizer, and in the bottom (Crafted I/Os) we use crafted I/Os to condition on the same synthesizer. These results show that using random I/Os on the same synthesizer leads to 15\%, and 5\% drops in the results for top-50 generalization and functional accuracy respectively. Based on these results we can see that random I/Os contain significantly less information about the target program in comparison to the crafted I/Os.

To further investigate the importance of the crafted I/Os we provide the results of synthesizing programs with different levels of complexity in \autoref{table:control}. In this table (\autoref{table:control}) we show the functional equivalency results of simple programs including programs that only contain action functions or \textit{Repeat} structure with action functions, and also complex programs that contain one or multiple conditional control flows.
\\
In \autoref{table:control} we can see that for simple programs that only contains action functions or action functions with \textit{Repeat} structure (Note that \textit{Repeat} is like for-loop structure, so any valid input can hit a \textit{Repeat} structure) we have low  performance drops in functional equivalence accuracy for Random I/Os in comparison to Crafted I/Os. For example, in the Action column (\autoref{table:control}) for top-50 accuracy, there is less than 1\% points drop for Random I/Os in comparison to the Crafted I/Os results. This is because any I/O examples can represent the functionality of these simple programs. In other words, any I/Os hits all parts of these simple programs. Furthermore, \autoref{table:control} shows that for more complex programs we have large drop in functional equivalence accuracy for Random I/Os in comparison to Crafted I/Os. As an example, we can see that for programs with \textit{While} structure in top-1 column, there is more than 27\% points drop for Random I/Os in comparison to the Crafted I/Os results. These results indicate that Crafted I/Os contain more informative details about the complex programs in comparison to random I/Os. This is because Crafted I/Os are designed to hit all branches of the programs. However, there is no guarantee if the given Random I/Os hit all of the branches of the complex programs.

\autoref{table:control} also provides the results of our iterative approach with and without fine-tuning the model. In this table (\autoref{table:control}) we can see that for the program with complex control flows our approaches have higher performance gain in comparison to the results for the simple program. This indicates that our iterative refinement approach is capable of generating more accurate programs by iteratively condition the model on informative I/Os. As an example, for the program with multiple control flows (Mix) in the top-1 column, we have around 17\% points improvement for Random I/Os~+~IReEn in comparison to Random I/Os.

\begin{figure}[h] 
	\centering
	\begin{subfigure}[b]{0.45\textwidth}
	    \centering
		\includegraphics[height=4.1cm]{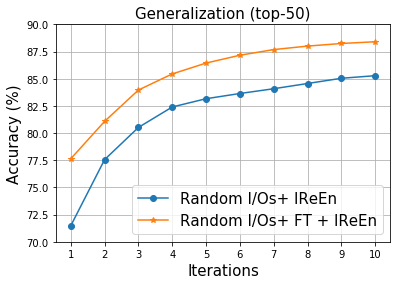}
		\caption{ }
		\label{fig:gentop50}
	\end{subfigure} 
	\hfill
	\begin{subfigure}[b]{0.45\textwidth}
	    \centering
		\includegraphics[height=4.1cm]{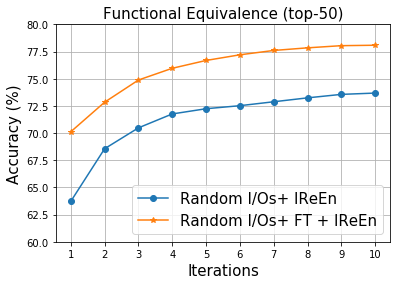} 
		\caption{}
		\label{fig:functop50}
	\end{subfigure} 
	\hfill
	\begin{subfigure}[b]{0.45\textwidth}
	    \centering
		\includegraphics[height=4.1cm]{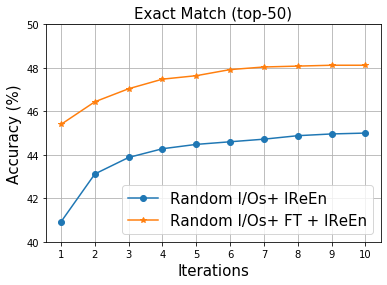} 
		\caption{}
		\label{fig:exacttop50}
	\end{subfigure} 
	\caption{(a) Generalization accuracy after each iteration for "Random I/Os + \,\ireng" and "Random I/Os + FT + \,\ireng". (b) Functional equivalence accuracy after each iteration for "Random I/Os+ \,\ireng" and "Random I/Os + FT + \,\ireng". (c) Exact match accuracy after each iteration for "Random I/Os + \,\ireng" and "Random I/Os + FT + \,\ireng". Note that, Random I/Os means that we use randomly obtained I/Os, FT denotes to the fine-tuned model, and \ireng\, refers to our iterative approach. }
	\label{fig:disc}
	\vspace{-0.4cm}
\end{figure}

\paragraph{Effectiveness of iterative refinement.} \autoref{fig:gentop50}, \autoref{fig:functop50}, and \autoref{fig:exacttop50} show the effectiveness of our proposed iterative approach in 10 iterations. In these figures, x and y axis refer to the number of iterations and the accuracy respectively. \autoref{fig:gentop50} shows the generalization accuracy for top-50, in \autoref{fig:functop50} we can see the results of functional equivalence metric for top-50, and \autoref{fig:exacttop50} demonstrates the exact match accuracy for top-50. In these figures, we provide results with and without fine-tuning the synthesizer. Here we can see the improvement of the generalization, functional equivalence, and exact match accuracy over the iterations. We have a margin of 7\% improvement in the functional equivalence accuracy for "Random I/Os + FT + \,\ireng" setting after 10 iterations. In other words, these results show that we can search for better random I/Os and program candidates by iteratively incorporate additional constraints. 

\begin{figure}[h] 
	\centering
	\begin{subfigure}[b]{0.45\textwidth}
	    \centering
		\includegraphics[height=4.1cm]{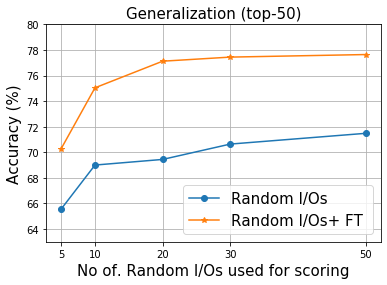}
		\caption{ }
		\label{fig:genscoring}
	\end{subfigure} 
	\hfill
	\begin{subfigure}[b]{0.45\textwidth}
	    \centering
		\includegraphics[height=4.1cm]{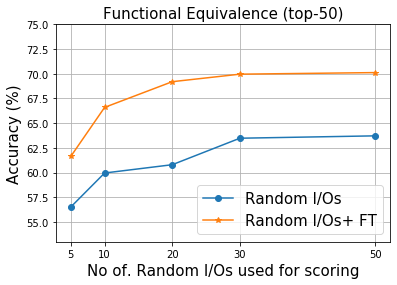} 
		\caption{}
		\label{fig:funcscoring}
	\end{subfigure} 
	\hfill
	\begin{subfigure}[b]{0.45\textwidth}
	    \centering
		\includegraphics[height=4.1cm]{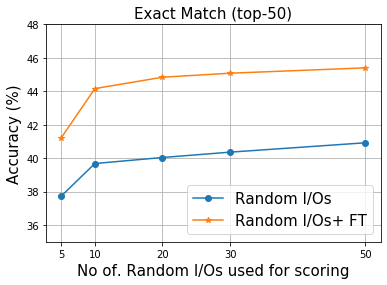} 
		\caption{}
		\label{fig:exactscoring}
	\end{subfigure}
	\caption{(a) Generalization accuracy after using the different numbers of random I/Os in scoring strategy for "Random I/Os" and "Random I/Os + FT". (b) Functional equivalence accuracy after using the different numbers of random I/Os in scoring strategy for "Random I/Os" and "Random I/Os + FT". (c) Exact match accuracy after using different numbers of random I/Os in scoring strategy for "Random I/Os" and "Random I/Os + FT". Note that, Random I/Os means that we use randomly obtained I/Os, FT denotes to the fine-tuned model, and \ireng\, refers to our iterative approach.}
	\label{fig:disc}
	\vspace{-0.4cm}
\end{figure}

\paragraph{Effectiveness of number of I/Os for sample rejection strategy.} In our approach in order to choose one candidate among all of the generated program candidates, we consider a sample rejection strategy. To do that, we use the random I/Os to assign a score to the generated candidates based on the number of satisfied random I/Os. Finally, we consider the candidate with the higher score as the best candidate and reject the rest. \autoref{fig:genscoring}, \autoref{fig:funcscoring}, and \autoref{fig:exactscoring} show the effect of using the different numbers of random I/Os on scoring the candidates and finding the best program candidate. x and y axis in these figures refer to the number of random I/Os, and accuracy of our approaches with and without fine-tuning. In \autoref{fig:genscoring} we show the generalization accuracy, in \autoref{fig:funcscoring} we provide functional equivalence results, and \autoref{fig:exactscoring} shows the results for exact match accuracy. These figures show that by using more random I/Os in the sample rejection strategy we can find more accurate programs that result to gain better performance in terms of generalization, functional equivalence, and exact match accuracy. In other words, with using more random I/Os for scoring the candidates we can capture more details of the black-box function, and find the best potential candidate among the generated one.

\section{Conclusion}

In this work, we propose an iterative neural program synthesis scheme to reverse-engineer the black-box functions and represent them in a high-level program. In contrast to previous works, where they have access to privileged information, in our problem setting, we only rely on the input-output interactions. To tackle the problem of reverse-engineering the black-box function in this challenging setting, we employ a neural program synthesizer in an iterative scheme. Using this iterative approach we search for the best program candidate in each iteration by conditioning the synthesizer on a set of violated constraints. Our evaluation on the Karel dataset demonstrates the effectiveness of our proposed approach in the reverse-engineering functional equivalent form of the black-box programs. Besides this, the provided results show that our proposed approach even outperforming the previous work that uses privileged information to sample input-output examples.

\clearpage
\bibliographystyle{splncs04}
\bibliography{ireen}

\end{document}